\documentclass[runningheads]{llncs}
\usepackage{graphicx}
\usepackage{bbding}
\usepackage{tikz}
\usepackage{comment}
\usepackage{amsmath,amssymb} 
\usepackage{color}
\usepackage{multirow}
\usepackage[accsupp]{axessibility}  
\usepackage[pagebackref=true,breaklinks=true,letterpaper=true,colorlinks,bookmarks=false]{hyperref}

\newcommand{\eg}{\textit{e}.\textit{g}.}

\newcommand{\figref}[1]{Fig.\ref{#1}}
\newcommand{\tabref}[1]{Tab.\ref{#1}}

\newcommand{\secref}[1]{Sec.\ref{#1}}
\newcommand{\etal}{\textit{et al}.}

\begin{document}
\pagestyle{headings}
\mainmatter
\def\ECCVSubNumber{935}  

\title{Occluded Human Body Capture with Self-Supervised Spatial-Temporal Motion Prior} 


\titlerunning{Occluded Human Body Capture}
%
\author{Buzhen Huang \and
Yuan Shu \and
Jingyi Ju \and
Yangang Wang}
\authorrunning{Huang et al.}
%
\institute{Southeast University, China
}
\maketitle

\begin{abstract}
    Although significant progress has been achieved on monocular maker-less human motion capture in recent years, it is still hard for state-of-the-art methods to obtain satisfactory results in occlusion scenarios. There are two main reasons: the one is that the occluded motion capture is inherently ambiguous as various 3D poses can map to the same 2D observations, which always results in an unreliable estimation. The other is that no sufficient occluded human data can be used for training a robust model. To address the obstacles, our key-idea is to employ non-occluded human data to learn a joint-level spatial-temporal motion prior for occluded human with a self-supervised strategy. To further reduce the gap between synthetic and real occlusion data, we build the first 3D occluded motion dataset~(OcMotion), which can be used for both training and testing. We encode the motions in 2D maps and synthesize occlusions on non-occluded data for the self-supervised training. A spatial-temporal layer is then designed to learn joint-level correlations. The learned prior reduces the ambiguities of occlusions and is robust to diverse occlusion types, which is then adopted to assist the occluded human motion capture. Experimental results show that our method can generate accurate and coherent human motions from occluded videos with good generalization ability and runtime efficiency. The dataset and code are publicly available at \url{https://github.com/boycehbz/CHOMP}.
\keywords{Human motion capture, Object occlusion, 3D human dataset, Spatial-temporal correlations}
\end{abstract}

\section{Introduction}
\label{sec:introduction}
Recovering 3D human motion from monocular images is a long-standing problem, which has wide applications such as computer animation, human behavior understanding, and human well-being. Recently, researches in this area have gained significant progress~\cite{pavllo20193d,luo20203d,kocabas2020vibe,zheng20213d,wang2017outdoor}, but most of them do not consider the occlusion scenarios that are very common in the real world.

Only a few works explicitly focus on 3D pose estimation from occluded images. Although~\cite{huang2009estimating,rafi2015semantic,sarandi2018robust,zhang2020object,kocabas2021pare,yang2022lasor,sun2021monocular,Pose2UV} can estimate 3D human poses from single occluded images, they do not utilize temporal information, and the results in highly ambiguous occlusion cases are unreliable. Some latest works~\cite{huang2021dynamic,rempe2021humor} incorporate the temporal information in a variational motion prior and can predict the full human motion from partial observations, but the optimization process in their methods is particularly time-consuming.

\begin{figure*}
    \begin{center}
    \includegraphics[width=1.0\linewidth]{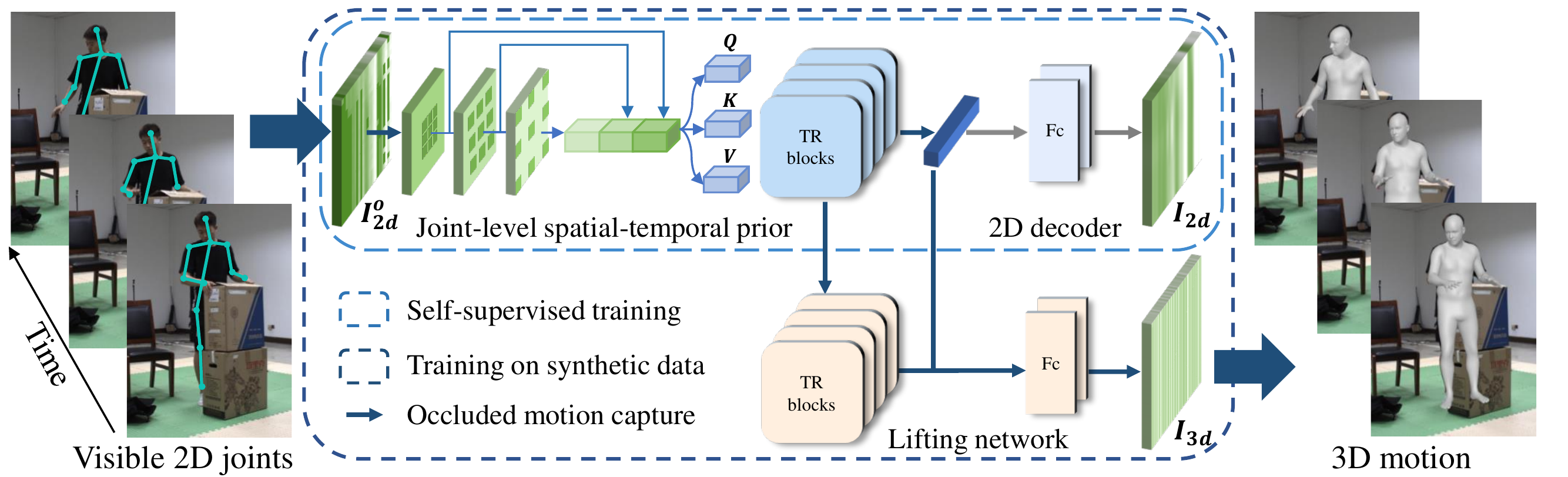}
    \end{center}
    \vspace{-8mm}
    \caption{The pipeline of our method. The 2D and 3D motions are represented with three 2D maps~($I_{2d}^o$, $I_{2d}$, and $I_{3d}$). We first train a joint-level spatial-temporal prior for occluded motions based on dilated convolutions and transformer blocks via self-supervised learning~(light blue box). The prior is then adopted to assist the lifting network for occluded human motion capture. The overall training can be conducted with synthetic occlusions on non-occluded data~(dark blue box). When testing on real occlusion data, the occluded map ($I_{2d}^o$) obtained with detectors is fed into the network to regress the 3D motion.}
    \label{fig:pipeline}
    \vspace{-6mm}
\end{figure*}

Training a neural network to regress 3D human motions from monocular occluded videos with temporal relations can significantly improve the runtime efficiency. However, the intuitive task faces two challenging obstacles. Due to occlusions and the loss of depth information, the problem is highly ambiguous as multiple 3D poses can map to the same 2D observations. On the other side, the limited amount of real occlusion 3D human data is still a bottleneck for training a robust temporal model. Thus, the networks cannot obtain stable and accurate results. Strong prior knowledge with compact motion representation is a feasible solution to address the challenges.

Our key-idea is \textbf{to employ non-occluded human data to learn a joint-level spatial-temporal prior for occluded human motion with a self-supervised strategy.} We then build a 3D occluded motion dataset~(\textbf{OcMotion}) to reduce the gap between synthetic and real occlusion data, which contains 43 motions and 300K frames with accurate 3D annotations. To the best of our knowledge, OcMotion is the first video-based dataset explicitly designed for the occlusion problem. To achieve the occluded motion capture, inspired by~\cite{zhang2020object} which represents a single occluded human in a 2D map, we propose an occluded motion map to simultaneously encode spatial and temporal information. As shown in~\figref{fig:representation}, with the map representation, we can synthesize occlusions on non-occluded data and eliminate the effect of image appearance. The intermediate representation also improves the human motion capture in generalization ability and accuracy~\cite{pavllo20193d,li2021exploiting,zhang2021learning}. Nevertheless, due to the loss of depth information and ambiguities of the occluded parts, it is difficult for the network to directly recover 3D human motion from the occluded 2D motion map. Thus, we learn a joint-level spatial-temporal prior to assist the 3D motion capture. We construct a 2D self-supervised task that regresses the full motion map from the synthetic occluded input. Different from the simple augmentation techniques~\cite{sarandi2018robust,biggs20203d,rockwell2020full,kocabas2021pare}, the self-supervised strategy recovers original 2D inputs from the occluded motion map, which formulates the problem as a masked image modeling~(MIM) task~\cite{bao2021beit,he2021masked} and can learn an expressive motion representation for the occluded problem. With the self-supervised training on a large amount of 2D data, the network also generalizes well on various occlusion types. To lift 3D human motion from the encoded motion features can alleviate the ambiguities induced by occlusions and produce more accurate results. In addition, previous video-based methods~\cite{kocabas2020vibe,choi2021beyond,pavllo20193d,li2021exploiting} all rely on image features~\cite{kolotouros2019learning} or pose features~\cite{li2021exploiting} to model the temporal relations among different frames, which ignore the joint-level kinematic information that is important for the occlusion problem. In contrast, with the map representation, we can further design a spatial-temporal layer based on dilated convolution and vision transformer~\cite{dosovitskiy2020image} to simultaneously consider the local joint-level correlations and global motion dependencies. The learned joint-level correlations can improve the prediction accuracy for occluded parts. In the inference phase, we detect the visible and reliable 2D joint coordinates with state-of-the-art detectors~\cite{cao2017realtime,fang2017rmpe,sun2019deep} to obtain the occluded 2D map, and then the full 3D human motion can be estimated with the trained model.

To sum up, the contributions of this paper are as follows:
\begin{itemize}
    \vspace{-1mm}
    \item We propose an occluded motion prior learned with self-supervised strategy on non-occluded data, which reduces the ambiguities of occluded observations and improves the motion capture in both generalization ability and accuracy.
    \item We propose a spatial-temporal layer with a map representation to simultaneously model local joint-level correlations and global motion dependencies for reliable occluded human motion capture.
    \item We build the first 3D occluded motion dataset, OcMotion, which contains 300K images captured in real occlusion scenarios. The dataset can be used for both training and testing. The dataset and code are publicly available. 
\end{itemize}

\section{Related Work}\label{sec:relatedwork}

\subsection{Occluded 3D human pose estimation}
Although the 3D human pose estimation has progressively developed in recent years, it still cannot achieve satisfactory performance in occlusion scenarios. Historically, there are few works~\cite{huang2009estimating,rafi2015semantic,sarandi2018robust,zhang2020object,kocabas2021pare,yang2022lasor} that explicitly focus on occluded human pose estimation. \cite{rafi2015semantic} relies on the depth information to infer the occluded body parts. To capture occluded human poses from RGB images, \cite{huang2009estimating} uses a linear combination of training samples, but it has limited expressive capabilities for various occlusion types and poses. Recently, ROMP~\cite{sun2021monocular,sun2021putting} regresses the occluded human from a single image by encoding the SMPL parameters in a 2D map, but it is not flexible enough to exploit temporal information. Other works~\cite{sarandi2018robust,biggs20203d,rockwell2020full,yang2022lasor,Pose2UV,wang2022best} also adopt neural networks for occluded pose estimation. However, no sufficient real occluded human data can be used to train a robust network, which has been the bottleneck of the regression-based methods for a long time. To improve occlusion-robustness, \cite{sarandi2018robust,biggs20203d,rockwell2020full} use synthetic occlusion data during training. To reduce the gap between synthetic and real occlusions, \cite{zhang2020object} represents the occluded human in the UV map and builds the first image-based object-occluded human dataset. Nonetheless, previous works that do not employ temporal information cannot obtain reliable results. Without the motion data for training, existing methods~\cite{rempe2021humor,huang2021dynamic} can only utilize the temporal information based on motion priors via a time-consuming optimization. In this work, we extend the dataset proposed by~\cite{zhang2020object} to have 43 real occluded 3D motions with complete and accurate annotations, thus we can train a temporal model for real-world occlusion problems. We also synthesize occlusions on existing non-occluded data to fully employ diverse motions. Different from the augmentation techniques in the previous methods~\cite{sarandi2018robust,biggs20203d,rockwell2020full,kocabas2021pare}, we propose a self-supervised strategy by recovering original 2D inputs from the occluded motion map, which can learn an expressive motion representation for the occluded problem via the MIM task~\cite{bao2021beit,he2021masked}. With the pre-trained motion prior, the model can regress full 3D motion from monocular occluded images with high accuracy and runtime efficiency.

\subsection{Human mesh recovery from monocular video}
Conventional methods~\cite{gall2009motion,wang2017outdoor,xu2018monoperfcap,arnab2019exploiting} fit predefined models to 2D image features to realize video-based human mesh recovery via solving a time-consuming constrained optimization. With the development of deep learning, temporal neural networks~\cite{kanazawa2019learning,sun2019human,liu2019temporally,zhao2021travelnet} are applied to learn the dependencies among frames. However, due to the limited training data, they use pseudo-ground-truth labels for training, which are unreliable for modeling accurate 3D human motion. \cite{kocabas2020vibe,luo20203d,choi2021beyond,yuan2022glamr} follow~\cite{kanazawa2019learning,sun2019human} to use the static features generated by image-based methods~\cite{kanazawa2018end,kolotouros2019learning} for modeling temporal relations. These methods depend heavily on the static features and ignore the kinematic information, which results in severe temporal inconsistency~\cite{kocabas2020vibe} and motion oversmoothness~\cite{choi2021beyond}. \cite{wan2021encoder} models the kinematic and temporal relations by attention mechanism, which has been proved useful for human mesh recovery from occluded inputs~\cite{wan2021encoder,chen2021self}. Although previous methods achieve competitive results on specific datasets, the generalization ability remains an obstacle due to the effect of image appearance and occlusions. Our method intermediately represents human motions in 2D maps and generalizes well to the change of environments. With the motion map, the local joint-level spatial-temporal correlations can be easily explored, which improves the human mesh recovery in occluded part inference.

\subsection{2D-to-3D lifting}
Due to the significant development of 2D pose detectors~\cite{cao2017realtime,chen2018cascaded,fang2017rmpe,sun2019deep}, 2D-to-3D lifting approaches generally outperform direct estimation methods in generalization ability and accuracy. Although many works~\cite{chen20173d,martinez2017simple,tekin2017learning,zhou2017towards,jiang20103d} achieve comparable results from single 2D poses, they can hardly be leveraged in real-world applications due to the loss of temporal dependencies. In recent years, some works~\cite{lee2018propagating,pavllo20193d,wang2020motion,li2021exploiting} add spatial-temporal information in 3D pose estimation, which greatly improve the accuracy and stability. \cite{lee2018propagating,hossain2018exploiting} rely on recurrent neural networks, but they cannot achieve parallel processing of multiple frames. Temporal convolution networks~\cite{pavllo20193d} with confidence heatmaps~\cite{cheng2019occlusion}, attention mechanism~\cite{liu2020attention} and body prior~\cite{chen2021anatomy} are also applied to explore spatial-temporal relations. Like graph convolutional networks~\cite{wang2020motion,cai2019exploiting}, dilated temporal convolutions are also inherently limited in temporal connectivity. Recently, transformer-based 3D human pose estimation~\cite{li2021exploiting,zheng20213d} show superior performance on long-term motions. However, they can only model information from one dimension. The joint-level spatial and temporal relations cannot be simultaneously considered, which is not suitable for the occlusion problem. Thus, we design a spatial-temporal layer with dilated convolutions to improve the current transformer-based structure in occluded human motion capture. The model can consider joint-level kinematic information and temporal dependencies in the same stage, which reduces ambiguities of occluded motions and consequently improves the accuracy and occlusion-robustness.

\section{Method}\label{sec:method}
Our goal is to recover human motion from monocular occluded images. We first represent the human motions in 2D maps and synthesize occluded motions using a large amount of non-occluded data~(\secref{sec:representation}). Since directly recovering 3D human motion from occluded 2D poses is a highly ambiguous problem, we then design a prior with a spatial-temporal layer to exploit joint-level correlations for occluded motions via self-supervised learning~(\secref{sec:prior}). The learned prior is then adopted to improve occluded 3D human motion capture~(\secref{sec:lifting}). Finally, we can obtain 3D human motion from an occluded video with 2D pose detectors in real-time~(\secref{sec:mocap}).

\subsection{Motion Representation}~\label{sec:representation}
To model joint-level spatial-temporal correlations for the occluded inputs, we represent the 2D motion and 3D motion in 2D maps. Different from previous map representations~\cite{sun2021monocular,zhang2020object} that encode body cues for single images, the motion map is more flexible in exploiting skeletal and temporal information. The 2D motion map $I_{2d}\in \mathcal{R} ^{F\times K\times 2}$ records the 2D joint coordinates of a motion sequence, which can be obtained by state-of-the-art 2D pose detectors~\cite{cao2017realtime,chen2018cascaded,fang2017rmpe,sun2019deep}. The $F$ and $K$ are frame length and number of joints. We normalize the 2D poses with the bounding-box for better generalization ability. The 3D motion map $I_{3d}\in \mathcal{R} ^{F\times N\times 6}$ is represented with SMPL parameters~\cite{loper2015smpl}. $N$ is the number of joints for SMPL. A 6D representation~\cite{zhou2019continuity} for the 3D rotation is adopted, thus the dimension of a joint rotation is 6. 

The motion map can be further used to represent the occluded motion by modifying the values in image pixels. Previous works~\cite{zhang2020object,zhang2021learning} represent the occluded joints with constant values~(\eg, 0) and inpaint all pixels of the target region. Although they can also obtain satisfactory 2D full maps via an inpainting task, we found the strategy cannot learn good representation for downstream tasks in self-supervised learning. Thus, we store a learned occlusion token in the pixel of an occluded joint in the motion map. In the inference phase, the occluded 2D motion map can be generated by replacing the detected low confidence joints with the learned token. With the occluded motion representation, we can train a joint-level spatial-temporal correlation prior for the occlusion problem and improve the generalization ability.

\begin{figure}
    \begin{center}
    \vspace{-4mm}
    \includegraphics[width=1.0\linewidth]{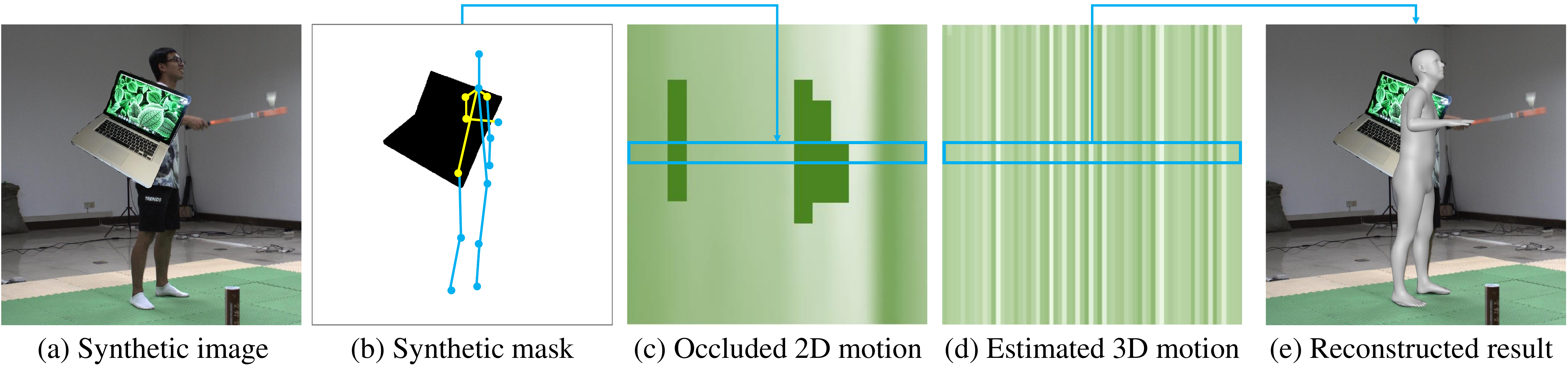}
    \end{center}
    \vspace{-8mm}
    \caption{We record 2D and 3D poses from consecutive frames in rows of 2D maps~(c,d). The dark green denotes occluded joints, which are represented with a learnable token. Since the human motion maps are not affected by image appearance, we can synthesize occlusions on a large amount of non-occluded data~(\eg, Human3.6M~\cite{ionescu2013human3}) to train a robust model.}
    \label{fig:representation}
    \vspace{-10mm}
\end{figure}

\subsection{Joint-level spatial-temporal correlation prior}~\label{sec:prior}
Due to the severe ambiguities and insufficient occlusion training data, it is challenging to learn an accurate and generalized model for the occluded human body capture. To address these obstacles, we use synthetic occlusion data to train a joint-level spatial-temporal correlation prior via self-supervised learning. The trained prior can learn the compact and expressive representation for the occluded 2D motion. The pre-trained network parameters with the learned prior knowledge are then used in the downstream lifting task to improve the occluded pose estimation.

\textbf{Occlusion data synthesis.} Since existing occluded human data is insufficient, we synthesize occlusions on non-occluded data~\cite{ionescu2013human3} and enforce the prior to exploit the essential motion information from the visible parts. To learn good representation from visible observations, recent works in masked image modeling tasks~\cite{bao2021beit,he2021masked} use random masks with a specific masking ratio to construct a self-supervised learning task. However, the strategy can hardly simulate real occlusion cases. Thus, we synthesize occlusions to generate an occluded 2D motion $I_{o2d}\in \mathcal{R} ^{F\times K\times 2}$ for the prior to learn more expressive occluded 2D motion representation. The procedures are shown in~\figref{fig:representation}~(a, b, c). We add a random occluder on the non-occluded image and store the visible 2D joint coordinates in the 2D motion map. Different from previous works~\cite{zhang2020object,zhang2021learning}, we use a learned occlusion token to represent the occluded joints in the map~(dark green region). The token promotes the network to learn better motion representation in self-supervised training. In addition, benefiting from our intermediate representation, the synthetic occlusions on the RGB images do not cause domain gap~\cite{zhang2020object}, thus we can learn a generalized prior with a large amount of data.

\textbf{Self-supervised training.} With the synthetic data, estimating accurate 3D human bodies from the occluded inputs is still a challenging problem. Previous works \cite{zhang2020object,kocabas2021pare,sarandi2018robust} apply a simple occlusion augmentation in 2D images and regress the 3D poses. However, directly recovering 3D motion from 2D poses itself is an ill-posed problem, and the occlusions further increase the uncertainty for the estimation, thus the same 2D observation can be mapped to various different 3D poses. They can still not obtain stable results for occluded cases. Thus, we propose a self-supervised strategy to learn a motion prior to reduce the ambiguities and fully exploit the motion information from visible cues. Different from previous works~\cite{zhang2020object,kocabas2021pare} that directly estimate 3D poses from 2D images, the self-supervised learning enforces the prior to recover the original full 2D map $I_{2d}$ from the occluded input $I_{o2d}$. With the self-supervised learning, the prior learns the expressive representation for the human motion~\cite{he2021masked,bao2021beit} with only the partial observations.

For the occluded pose estimation, the kinematic information and temporal dependencies are bases to infer an occluded body part. The joint-level spatial-temporal correlations are essential cues. Conventional methods~\cite{kocabas2020vibe,choi2021beyond,pavllo20193d,li2021exploiting} model the temporal relations among different frames based on image or pose features, which ignore the local kinematic information. Although a recent work~\cite{zheng20213d} considers both the spatial and temporal aspects with distinct transformer modules~\cite{dosovitskiy2020image}, it explores spatial and temporal relations in different stages. The temporal information cannot be considered in the spatial stage, thus some joint-level temporal features may be lost. Therefore, we design a dilated convolutional layer to extract local joint-level spatial-temporal features from the occluded 2D map. As shown in~\figref{fig:pipeline}, the joint-level spatial-temporal layer consists of 3 convolutional layers with different dilations. The convolutional kernel can simultaneously model joint relations from the kinematic structure and temporal features, which is essential for inferring occluded joints.

Since the dilated convolutional layers are limited in temporal connectivity, we concatenate the 3 output feature maps from the joint-level spatial-temporal layer and use two stacked transformers~\cite{dosovitskiy2020image} to model global information. The concatenated map $f\in \mathcal{R} ^{F\times K\times D}$ has the same resolution as the occluded map, and $D$ is the dimension of the vectors in feature map pixels. We then add a learnable spatial positional embedding $E_{SPos}\in \mathcal{R} ^{K\times D}$ in the skeletal dimension of the feature map. The resulting skeletal features are fed into the first transformer encoder to exploit information across the skeletal structure. We flatten the output features to $f_T\in \mathcal{R} ^{F\times (K \cdot D)}$ and add a learnable temporal positional embedding $E_{TPos}\in \mathcal{R} ^{F\times (K \cdot D)}$. The features are then encoded with the second transformer module. Finally, a regression head with an MLP block and a Layer norm is used to obtain a full 2D motion map $I_{2d}$.

As shown in the light blue box in \figref{fig:pipeline}, the training does not require extra 3D annotations, and we can employ diverse 2D motion data~\cite{andriluka2018posetrack,zhang2013actemes} to learn a robust prior. A large amount of synthetic data also relieves the data-hungry problem for the transformer-based prior in the occluded pose estimation.

\textbf{Loss function.} We use the following loss function for the self-supervised training:
\begin{equation}
    \mathcal{L}_{self}=M_{2d}\left\| I_{2d}-I_{2d}^{gt}\right\|_1,
\end{equation}
The L1 loss is applied to the masked region of the predicted motion map. $M_{2d}$ is the synthetic mask, where the occluded part is 1, and the rest are 0. We found that the L1 loss without any other smoothness term can promote the prior to learn better motion representation in the self-supervised task.

When the training is completed, we use the pre-trained prior in the lifting task to assist the 3D motion capture.

\subsection{Occluded motion lifting}~\label{sec:lifting}
The encoder of the trained prior is then used in the lifting network to improve the occluded motion estimation. The lifting network is a transformer with two heads for 3D motion map and shape parameters. The transformer module has the same structure as the first transformer in the prior and receives the feature map $f$ from the convolutional layer. We then add the output features to the features from the prior and feed them to the heads to obtain 3D motion map $I_{3d}$ and shape parameters $\beta$. With the learned motion representation, the lifting network can achieve better accuracy and generalization, and we also optimize the network parameters of the prior in the lifting stage. Since the 2D occluded motion maps are not affected by image appearance, the training of the lifting network can also be conducted on synthetic non-occluded data~\cite{ionescu2013human3,sarandi2018robust}. The loss function is:
\begin{equation}
    \mathcal{L}_{motion}= \mathcal{L}_{rec} + \mathcal{L}_{shape} + \mathcal{L}_{smo}.
\end{equation}
The reconstruction term is:
\begin{equation}
    \mathcal{L}_{rec}= \left\| I_{3d}-I_{3d}^{gt}\right\|_2 + \left\| V_{3d}-V_{3d}^{gt}\right\|_2 + \left\| J_{3d}-J_{3d}^{gt}\right\|_2.
\end{equation}

The $I_{3d}$, $V_{3d}$ and $J_{3d}$ are predicted 3D motion map, vertex positions and joint positions. $gt$ denotes the ground-truth. To prevent the jitters and obtain coherent motions, we further add a smoothness loss on the predicted 3D motion map.
\begin{equation}
    \mathcal{L}_{smo}= \sum_{i=0}^{F-1} \left\| I_{3d}^i - I_{3d}^{i+1} \right\|_2,
\end{equation}

where $i$ denotes the $i$th row of $I_{3d}$. Additional shape and regularization terms are applied to prevent abnormal body shape:
\begin{equation}
    \mathcal{L}_{shape}=  \left\| \beta-\beta^{gt}\right\|_2 + \left\| \beta\right\|_2.
\end{equation}

\subsection{Occluded motion estimation from monocular videos}~\label{sec:mocap}
With the trained spatial-temporal motion prior and the lifting network, we can recover full 3D motion from occluded images. We first adopt~\cite{cao2017realtime} to detect 2D poses and corresponding confidence maps. The joint with a confidence lower than 0.6 will be regarded as the occluded joint, which will be replaced with the learned token. Thus, the occluded 2D map can be obtained. We then regress the full 3D motion map through the prior and lifting network from the occluded 2D map. Since the 3D motion resampled from the map is in the local coordinate system, we obtain global positions by solving the least square function.
\begin{equation}
    \mathop{\arg\min}_{\left(\mathcal{T}\right)_{0: F-1}} \mathcal{L}=\sum^{F-1}_{t=0}{ (w_c \left\| K(J_{3d}^t+\mathcal{T}^t)-P^{t}\right\| + \left\| \mathcal{T}^{t+1} - \mathcal{T}^{t} \right\| )},
\end{equation}
where $K$ is intrinsic camera parameters, $ P$ and $w_c$ are detected 2D pose and corresponding confidence. $\mathcal{T}$ is the translation of the SMPL model. Finally, the 3D motion with absolute positions can be obtained by adding the translations to the estimated SMPL models.

\section{OcMotion Dataset}\label{sec:Dataset}

\begin{figure*}
    \vspace{-8mm}
    \begin{center}
    \includegraphics[width=1.0\linewidth]{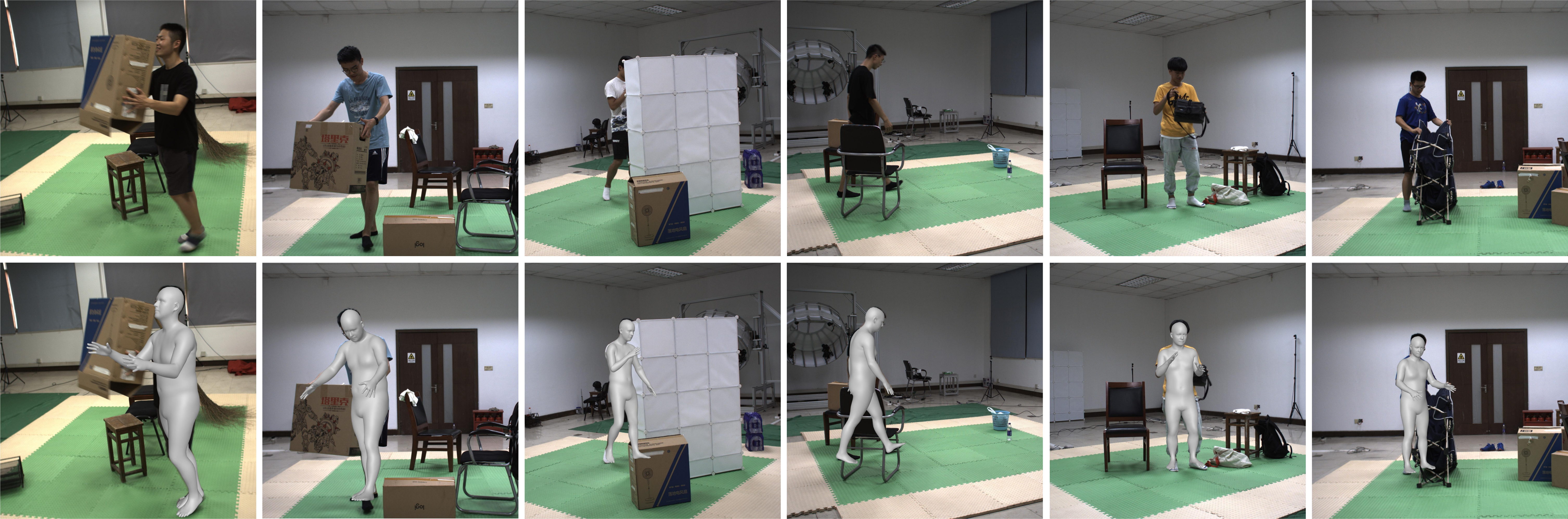}
    \end{center}
    \vspace{-8mm}
    \caption{Samples from the proposed OcMotion dataset. The dataset contains 300K images captured at 10 FPS~(frame per second) with accurate 3D motion annotations.}\label{fig:dataset}
    \vspace{-8mm}
\end{figure*}

Although 3D human datasets are exponentially increasing in recent years, only a few datasets are particularly designed for occlusion problems. AGORA~\cite{patel2021agora} contains frequent occlusions, but it is a synthetic dataset. 3DOH50K~\cite{zhang2020object} is the first 3D human dataset that explicitly considers object occlusion. However, it is an image dataset and cannot be used for evaluating video-based methods. In this work, we extend the 3DOH50K dataset to have complete motion annotations. We obtain human motions with~\cite{huang2021dynamic} and the samples in 3DOH50K are also used for constraining the optimization. For severely occluded cases, we manually adjust the 3D motion. To evaluate the accuracy of the dataset, we randomly select 5K images and manually annotate 2D poses. The re-projection error on these images is 7.3 pixels, which is sufficient for motion capture tasks. Finally, the dataset has 300K images captured at 10 FPS, 43 sequences with 6 viewpoints, 3D motion annotations represented by SMPL, 2D poses, and camera parameters.

\begin{table}
    \caption{Comparison with commonly used 3D human datasets. OcMotion is the first motion dataset that contains diverse real object occlusions with complete and accurate annotations.}
    \label{tab:dataset}
    \vspace{-5mm}
    \begin{center}
        \resizebox{1.0\linewidth}{!}{
            \begin{tabular}{l|c|c|c|c|c|r|r}
            \noalign{\hrule height 1.5pt}
            Dataset &Occlusion Data &Sequence &Real Data &3D Pose &Mesh &Frames &Views \\
            \noalign{\hrule height 1pt}
            Human3.6M~\cite{ionescu2013human3}      &-- &\checkmark &\checkmark &\checkmark &-- &3.6M &4 \\ 
            AIST++~\cite{li2021ai}         &-- &\checkmark &\checkmark &\checkmark  &\checkmark &10.1M &9  \\   
            HUMBI~\cite{yu2020humbi}          &-- &\checkmark &\checkmark &\checkmark  &\checkmark &17.3M &107 \\
            MPI-INF-3DHP~\cite{mehta2017monocular}   &+ &\checkmark &\checkmark &\checkmark  &-- &1.3M &14\\
            3DPW~\cite{von2018recovering}           &++  &\checkmark &\checkmark &\checkmark &\checkmark &50K &1 \\ 
            MuPoTs-3D~\cite{mehta2018single}      &++ &\checkmark &\checkmark &\checkmark   &-- &8K &1\\
            Panoptic Studio~\cite{joo2017panoptic} &++ &\checkmark &\checkmark &\checkmark  &-- &1.5M &480 \\
            GPA~\cite{wang2019geometric}           &++ &\checkmark &\checkmark &\checkmark  &-- &700K &5 \\
            3DOH50K~\cite{zhang2020object}        &++++  &-- &\checkmark &\checkmark      &\checkmark &50K &6 \\
            AGORA~\cite{patel2021agora}          &++++ &-- &-- &\checkmark  &\checkmark &17K &-- \\
            \textbf{OcMotion} &++++ &\checkmark &\checkmark &\checkmark  &\checkmark &300K &6\\
            \noalign{\hrule height 1.5pt} 
            \end{tabular}
        }
    \end{center}
    \vspace{-10mm}
    \end{table}
    
We show the comparison with commonly used 3D human datasets in~\tabref{tab:dataset}. Despite a large number of samples or the wide variety of actions, most existing datasets pay little attention to the occlusion problem. MPI-INF-3DHP~\cite{mehta2017monocular}, MuPoTs-3D~\cite{mehta2018single} and Panoptic Studio~\cite{joo2017panoptic} include only a few occluded cases. GPA~\cite{wang2019geometric} has limited types of occluders. In addition, AGORA~\cite{patel2021agora} and 3DOH50K~\cite{zhang2020object} are image-based datasets and cannot be applied to video-based methods. In contrast, our dataset contains complete motion annotations and explicitly considers object-occluded scenarios, which may promote future research on object-occluded human mesh recovery.

\section{Experiments}
\label{sec:experiments}

\subsection{Implementation details}
The spatial-temporal layer consists of 3 dilated convolution layers with a dilation rate of 1, 2, and 5, respectively. We also adopt two separate transformers~\cite{dosovitskiy2020image} to model long-term spatial and temporal information. The decoder of the 2D branch only contains a LayerNorm~\cite{ba2016layer} and a linear layer, thus the features encoded by the prior can represent a full motion. The lifting network is also a transformer. We rely on PyTorch~\cite{paszke2019pytorch} to implement the model and use AdamW~\cite{loshchilov2017decoupled} optimizer with a learning rate of 1e-4 for training. The batchsize of all experiments is 32. The model is trained on a single NVIDIA RTX 3090 GPU with 24GB memory for 45 epochs. We use a joint regressor in LSP format~\cite{johnson2010clustered} to obtain 3D joints and calculate errors for the predicted mesh.

\begin{table}
    \caption{Quantitative comparison with state-of-the-art methods. Our method obtains good results and achieves the best performance in some metrics on occluded datasets. $^\ast$ means the image-based method. $^\dagger$ denotes the method that explicitly considers the occlusion problem. Total Params is the total number of model parameters including 2D detection~\cite{cao2017realtime} or feature extraction~\cite{kolotouros2019learning}.}
    \label{tab:experiment}
    \vspace{-3mm}
    \begin{center}
        \resizebox{1.0\linewidth}{!}{
            \begin{tabular}{l|c|c c c|c c c| c c c}
            \noalign{\hrule height 1.5pt}
            \begin{tabular}[l]{l}\multirow{2}{*}{Method}\end{tabular} &\begin{tabular}[l]{l}\multirow{2}{*}{\begin{tabular}[1]{c}Total\\Params\end{tabular}}\end{tabular} &\multicolumn{3}{c|}{OcMotion} &\multicolumn{3}{c|}{3DPW}  &\multicolumn{3}{c}{3DPW-OC}\\
            & &MPJPE &PA-MPJPE &Accel. &MPJPE &PA-MPJPE &PVE &MPJPE &PA-MPJPE &Accel.\\
            \noalign{\hrule height 1pt}
            HMMR~\cite{kanazawa2019learning}    &29.8M  &-- &-- &-- &116.5 &72.6 &139.3 &-- &-- &--   \\
            $^\ast$SPIN~\cite{kolotouros2019learning}    &27.0M   &88.2 &56.7 &47.0 &96.9 &59.2 &116.4 &105.0 &71.3 &44.6   \\   
            $^\ast$$^\dagger$OCHMR~\cite{khirodkar2022occluded}  &-- &--  &-- &-- &89.7 &58.3  &107.1  &-- &-- &--\\
            $^\ast$$^\dagger$LASOR~\cite{yang2022lasor} &-- &-- &-- &-- &-- &57.9 &-- &-- &-- &--  \\
            VIBE~\cite{kocabas2020vibe}         &48.3M  &89.6 &58.6 &44.5 &93.5 &56.5 &113.4 &98.3 &69.7 &39.0   \\ 
            TCMR~\cite{choi2021beyond}     &108.9M  &95.8  &62.6  &24.3  &95.0 &55.8 &111.5 &90.3 &63.0 &\textbf{8.0}   \\
            $^\ast$$^\dagger$OOH~\cite{zhang2020object} &33.0M  &83.0 &55.0 &48.6 &86.7 &55.2 &105.2 &90.4 &57.0 &45.3 \\
            MEVA~\cite{luo20203d}     &92.0M &88.8 &59.9 &29.0 &86.9 &54.7 &-- &91.4 &63.5 &17.8   \\
            $^\ast$$^\dagger$ROMP~\cite{sun2021monocular}  &29.0M &79.4  &48.1 &57.2 &85.5  &53.3  &103.1 &--  &66.5  &--\\
            $^\ast$$^\dagger$PARE~\cite{kocabas2021pare}   &32.9M   &81.1 &52.0 &43.6 &\textbf{82.9} &52.3 &\textbf{99.7} &90.5 &56.6 &40.9 \\
            Wan~\etal~\cite{wan2021encoder}     &-- &-- &-- &-- &88.8 &50.7 &104.5 &-- &-- &--   \\ 
            Chen~\etal~\cite{chen2021self}     &51.4M &-- &-- &-- &85.8 &\textbf{50.4} &100.6 &-- &-- &--   \\ 
            \textbf{$^\dagger$Ours}~(w/o OcMotion) &56.3M &72.1 &44.9 &24.2 &83.7 &51.8 &110.4 &90.1 &54.5 &16.6 \\
            \textbf{$^\dagger$Ours} &56.3M &\textbf{58.3} &\textbf{36.1} &\textbf{23.2} &83.7 &51.7 &110.1 &\textbf{89.4} &\textbf{53.4} &16.6 \\
            \noalign{\hrule height 1.5pt} 
            \end{tabular}
        }
\end{center}
\vspace{-10mm}
\end{table}

\noindent\textbf{Metrics}. We adopt the metrics in previous works~\cite{kocabas2020vibe,luo20203d,choi2021beyond} to evaluate our method. The Mean Per Joint Position Error (MPJPE) and the MPJPE after rigid alignment of the prediction with ground truth using Procrustes Analysis (PA-MPJPE) are used for measuring joint positions. The Per Vertex Error (PVE) and acceleration error (Accel.) are applied to evaluate mesh quality and motion smoothness.

\subsection{Dataset}
We follow previous works~\cite{kocabas2020vibe,choi2021beyond} to set the training data for a fair comparison. The spatial-temporal prior is trained on 2D~(PoseTrack~\cite{andriluka2018posetrack}, InstaVariety~\cite{kanazawa2019learning} and PennAction~\cite{zhang2013actemes}) and 3D human motion datasets~(Human3.6M~\cite{ionescu2013human3} and MPI-INF-3DHP~\cite{mehta2017monocular}). We use Human3.6M and MPI-INF-3DHP to train the lifting network. The method is evaluated on 3DPW~\cite{von2018recovering}, Human3.6M, and MPI-INF-3DHP. In addition, we use 3DPW-OC, the occluded sequences from the entire dataset selected by~\cite{zhang2020object}, to demonstrate the effectiveness of our approach in occluded cases. We also report the results with and without OcMotion training. The sequences \textit{0013, 0015, 0017, 0019} in OcMotion with 6 views are used for testing, and the rest are adopted for training.

\begin{figure*}
    \begin{center}
    \includegraphics[width=1\linewidth]{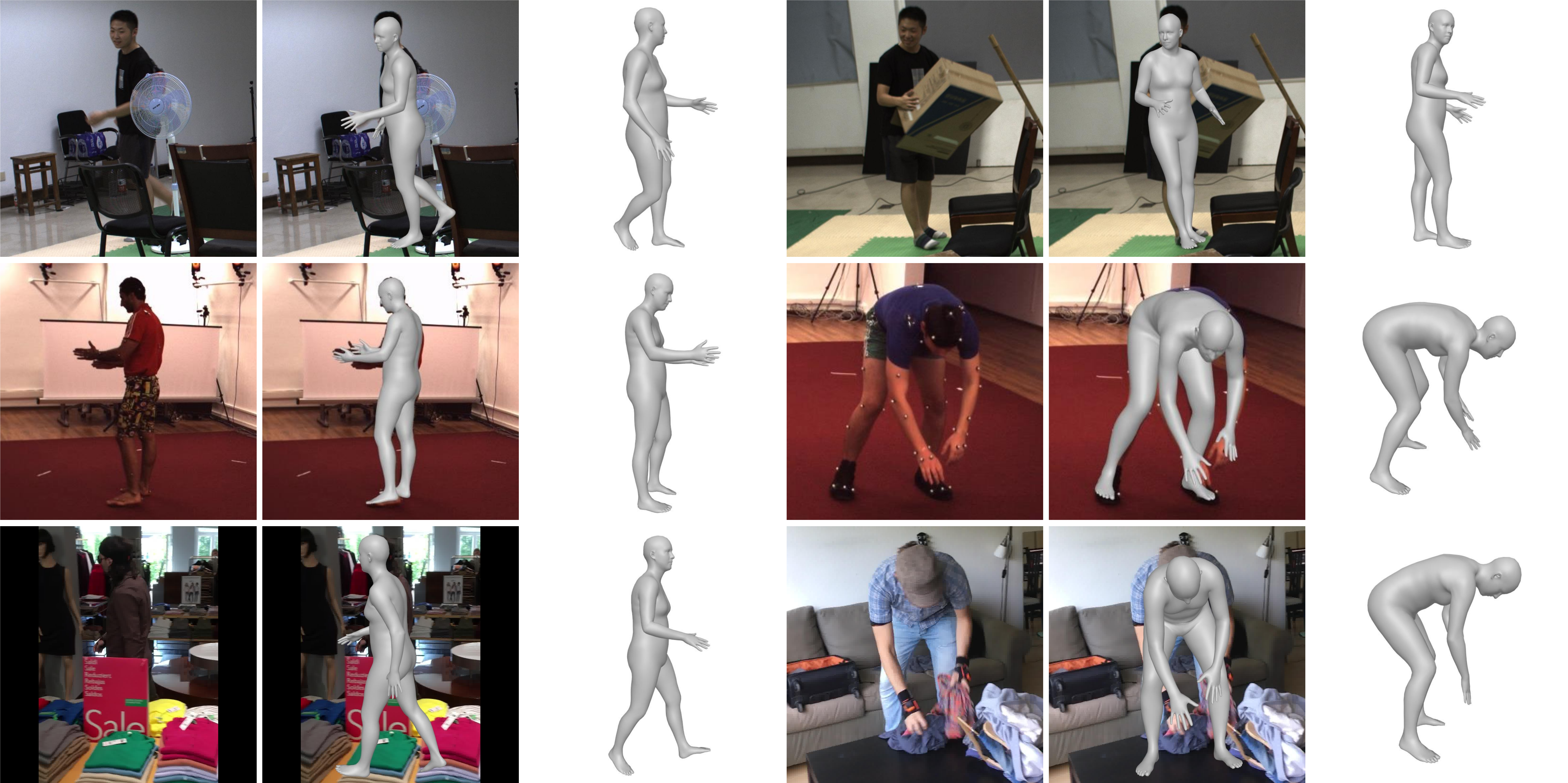}
    \end{center}
    \vspace{-6mm}
    \caption{Our method achieves good performance in both occluded and non-occluded cases.}
    \label{fig:quali_results}
    \vspace{-7mm}
\end{figure*}

\subsection{Comparison to state-of-the-art results}
\begin{figure*}
    \vspace{-4mm}
    \begin{center}
    \includegraphics[width=1\linewidth]{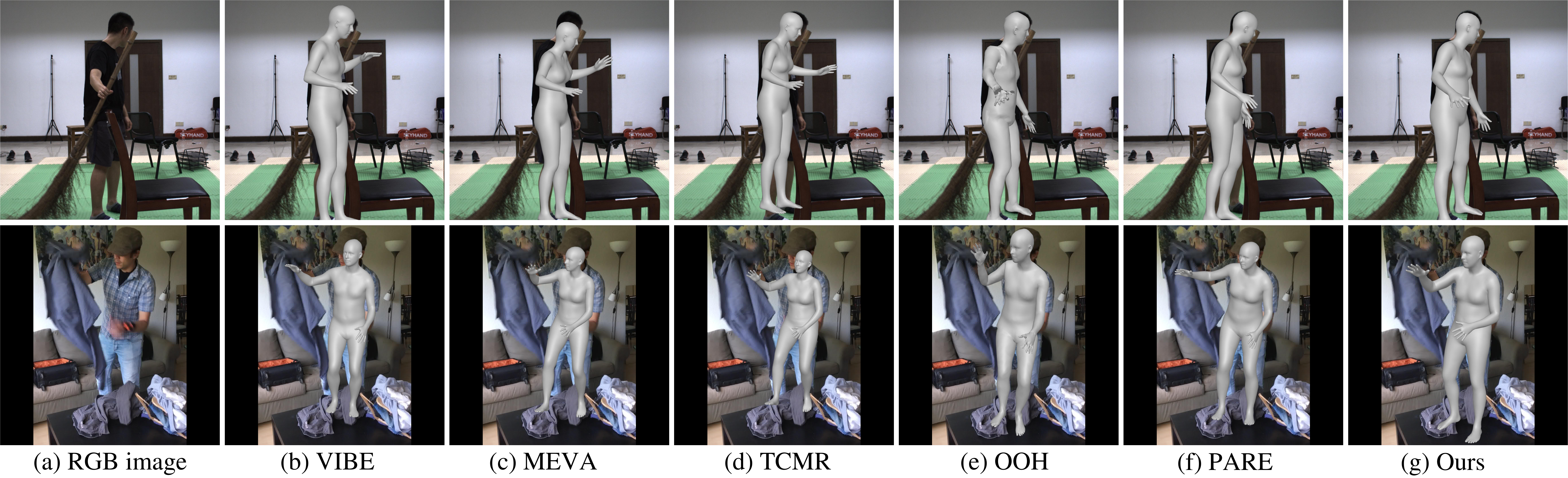}
    \end{center}
    \vspace{-8mm}
    \caption{Qualitative comparison among the methods that utilize temporal information~(b, c, d) and explicitly consider the occlusion problem~(e, f). Our method is more robust to occlusions than other methods.}
    \label{fig:results}
    \vspace{-8mm}
\end{figure*}

\begin{figure*}
    \vspace{-6mm}
    \begin{center}
    \includegraphics[width=1\linewidth]{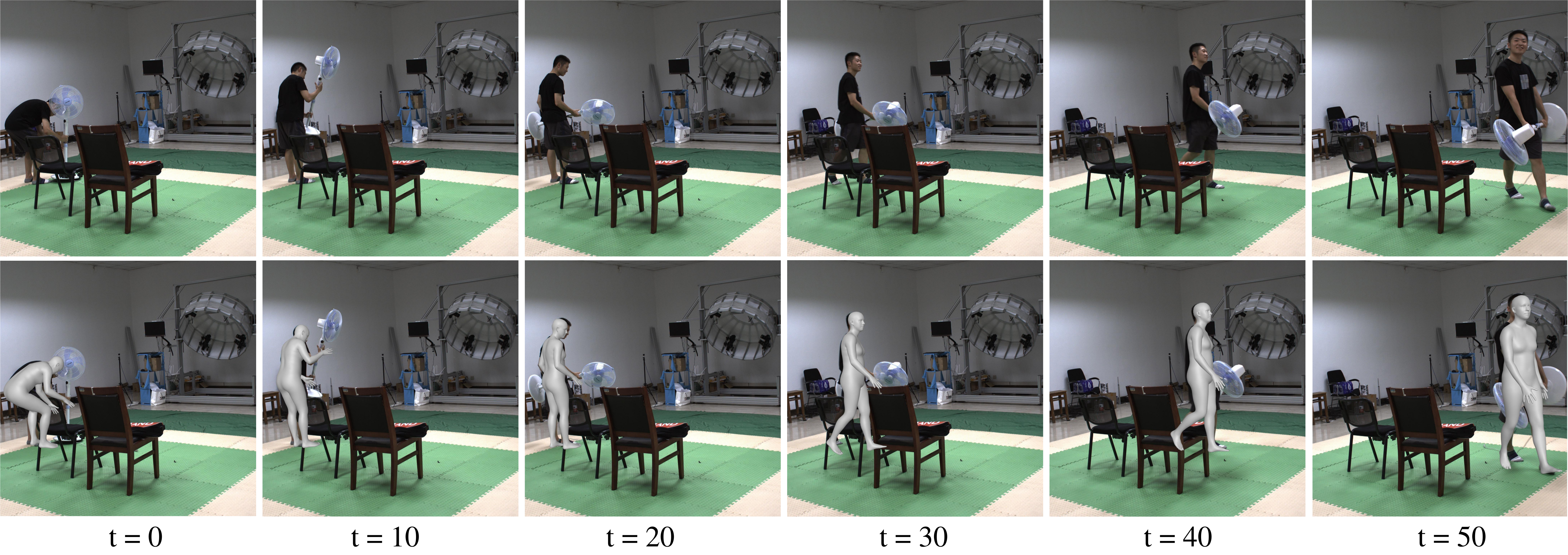}
    \end{center}
    \vspace{-8mm}
    \caption{Qualitative results on consecutive frames in occlusion scenario. More results can refer to our supplemental video.}
    \label{fig:qualitative_video}
    \vspace{-8mm}
\end{figure*}

We first compared our method to state-of-the-art approaches on OcMotion dataset. To the best of our knowledge, OcMotion is the first video dataset designed for the object-occluded human mesh recovery task. We conducted experiments on this dataset to demonstrate the superiority of our method in occluded cases. As shown in~\tabref{tab:experiment}, since previous methods do not explicitly consider the occlusion problem, our method significantly outperforms previous video-based methods on all metrics. In addition, PARE~\cite{kocabas2021pare} and OOH~\cite{zhang2020object} are image-based methods and are designed for the occluded human reconstruction task. Benefited from the spatial-temporal information, our method can obtain more robust results. Moreover, OOH~\cite{zhang2020object} uses UV map representation, though it can explicitly describe an occluded human, the resampled meshes show a lot of artifacts~\figref{fig:results}~(e). Furthermore, we found that video-based methods show more motion jitters and have higher acceleration errors in occluded cases. In contrast, the spatial-temporal prior obtains features of a full motion and avoids the ambiguities induced by occlusions. With the prior, our method achieves the best performance in terms of acceleration error and produces more temporally coherent results~\figref{fig:qualitative_video} on the occlusion dataset.

Another strength of our method is the good generalization ability. Since we represent the human motion in a 2D map, the background of the image cannot affect the model performance, thus the trained model is insensitive to environmental changes. Besides, the self-supervised training on a large amount of synthetic data makes the model robust to various occlusions. We conduct some experiments on 3DPW dataset to demonstrate the advantages. Our model is trained on indoor 3D datasets but can also achieve satisfactory performance in outdoor scenarios. Specifically, \cite{wan2021encoder} and \cite{chen2021self} also develop attention modules to exploit temporal cues and produce the best results. The results in~\tabref{tab:experiment} demonstrate that our method can also obtain similar results as state-of-the-arts on the non-occluded dataset. To further show the performance in more challenging occluded environments, we follow~\cite{zhang2020object} to evaluate the method on 3DPW-OC dataset, which is a subset of 3DPW that contains occlusions. We have the same training data as VIBE, MEVA, and TCMR. Although VIBE, MEVA, and TCMR achieve excellent results on 3DPW, they are sensitive to the occlusion~\figref{fig:results}~(b,c,d), while our method is robust. Moreover, TCMR relies on the past and future features to regress SMPL parameters for the current frame, and it may produce over-smoothed motion~(refer to supplemental video). With the spatial-temporal prior, our method can obtain coherent and highly dynamic motions.

We conduct experiments on Human3.6M dataset to further demonstrate the effectiveness of our method. As shown in~\tabref{tab:h36m}, our method achieves state-of-the-art in protocol 1 of Human3.6M in terms of MPJPE, which outperforms VIBE by 4.1mm. In addition, in~\tabref{tab:experiment} and~\tabref{tab:h36m}, LASOR~\cite{yang2022lasor} and Pose2Mesh~\cite{choi2020pose2mesh} also recover human mesh from 2D detections, and our method can obtain more accurate results.

We also report the number of model parameters to show the efficiency of our approach. The results in~\tabref{tab:experiment} show the total parameters including 2D pose detection~\cite{cao2017realtime} and feature extraction~\cite{kolotouros2019learning} of different methods. We have fewer parameters than TCMR and MEVA. Specifically, the lifting network in our method contains 19.2M parameters. The 2D detection consumes most of the computations, which can be replaced with more compact models. To further demonstrate the runtime efficiency, we report the model parameters without the 2D detection and feature extraction among the two-stage methods in~\tabref{tab:h36m}. Our model can achieve the best performance with the fewest parameters. For current implementation with~\cite{cao2017realtime}, the inference FPS is 32 on a single NVIDIA RTX 3090 GPU, which is acceptable for real-time applications. Furthermore, when regressing from the 2D pose inputs, our model runs at 286 FPS, which is significantly faster than common 2D pose detectors. The inference speed will not be the bottleneck for real-world implementation.

\begin{table}
    \vspace{-5mm}
    \caption{We conduct quantitative comparisons on Human3.6M following the protocols defined in~\cite{kanazawa2018end}. In the column of \#Params, we also compare the number of model parameters without 2D detection and feature extraction among two-stage methods. MACs is the estimated multiply–accumulate operations. Our method achieves competitive performance as state-of-the-art methods with higher runtime efficiency.}
    \label{tab:h36m}
    \vspace{-2mm}
    \begin{center}
        \resizebox{0.65\linewidth}{!}{
            \begin{tabular}{l|c|c|c c|c c c}
            \noalign{\hrule height 1.5pt}
            \begin{tabular}[l]{l}\multirow{2}{*}{Method}\end{tabular} &\begin{tabular}[l]{l}\multirow{2}{*}{\#Params}\end{tabular} &\begin{tabular}[l]{l}\multirow{2}{*}{MACs}\end{tabular} &\multicolumn{2}{c|}{Protocol 1} &\multicolumn{3}{c}{Protocol 2}\\
            & & &MPJPE &PA-MPJPE &MPJPE &PA-MPJPE &Accel.\\
            \noalign{\hrule height 1pt}
            HMMR~\cite{kanazawa2019learning}    &-- &--  &-- &-- &-- &56.9 &--\\
            MEVA~\cite{luo20203d}    &65.0M &1.31G   &73.4 &51.9 &76.0 &53.2  &15.3\\
            $^\ast$$^\dagger$PARE~\cite{kocabas2021pare}  &-- &--  &78.5 &55.1   &71.6 &49.9 &32.3         \\
            $^\ast$Pose2Mesh~\cite{choi2020pose2mesh}   &76.3M &3.81G   &-- &--   &64.9 &47.0 &--         \\
            DSD-SATN~\cite{sun2019human}  &-- &--  &--  &--  &59.1  &42.4 &--\\
            VIBE~\cite{kocabas2020vibe}   &21.3M &0.45G   &68.8 &49.5   &65.9 &41.5 &27.3  \\
            $^\ast$$^\dagger$OOH~\cite{zhang2020object} &-- &--  &74.7 &53.3  &61.8 &41.2 &35.3  \\
            TCMR~\cite{choi2021beyond}     &81.9M  &1.29G   &79.8 &56.8  &62.3  &41.1 &\textbf{5.3} \\
            Chen~\etal~\cite{chen2021self}    &-- &--  &-- &-- &58.9 &\textbf{38.7} &--\\
            Wan~\etal~\cite{wan2021encoder}     &-- &--  &-- &-- &\textbf{56.3} &\textbf{38.7} &--\\
            \textbf{$^\dagger$Ours}~(w/o OcMotion) &\textbf{19.2M} &0.49G    &\textbf{64.7} &\textbf{46.3} &59.7 &40.1 &13.0 \\
            \noalign{\hrule height 1.5pt} 
            \end{tabular}
        }
\end{center}
\vspace{-10mm}
\end{table}

\subsection{Ablation}

\noindent\textbf{Self-supervised prior.}
We ablate the self-supervised spatial-temporal prior to reveal the properties of this module in occluded human motion capture. We found that the occlusion token can promote the prior to learn better motion representation in self-supervised learning. In \tabref{tab:Ablation}, we remove the token and use constant value 0 to represent the occluded parts as~\cite{zhang2020object}. Using the prior learned with the constant value even produces a worse performance. The experimental results demonstrate the importance of the occlusion token. We then compared the model with and without the prior. The two strategies also use the occlusion augmentation technique~\cite{sarandi2018robust}. The results show that the prior significantly improves the joint and vertex accuracy. Directly recovering 3D human motion from occluded 2D poses without the prior is a highly ill-posed problem, and the same occluded 2D pose can map to various 3D poses. However, with the self-supervised training on many 2D motions and synthetic occlusions, the model learns the prior knowledge of converting an occluded motion to a highly dynamic full motion. The results show that with the assistance of the learned prior, the occluded human motion capture can obtain more accurate and coherent motions.

\begin{table}
    \vspace{-6mm}
    \caption{Ablation studies on different key components on OcMotion dataset. The spatial-temporal layer~(ST layer) and the self-supervised prior improve the motion capture in both joint accuracy and motion smoothness in occluded cases. + denotes adding the corresponding module on the temporal model.}
    \label{tab:Ablation}
    \vspace{-3mm}
    \begin{center}
        \resizebox{1.0\linewidth}{!}{
            \begin{tabular}{l|c|c|c c c c}
            \noalign{\hrule height 1.5pt}
            Method &\#Params &MACs &MPJPE&PA-MPJPE &PVE &Accel.\\
            \noalign{\hrule height 1pt}
            temporal &19.1M &0.46G &70.5 &44.8 &79.6 &40.6 \\
            + smoothness loss &19.1M &0.46G &72.6  &45.4   &80.1   &27.4  \\
            + spatial &19.1M &0.47G &65.4  &41.1   &76.4   &40.4  \\
            + ST layer &19.2M &0.49G &63.6 &39.8 &73.7 &38.6  \\
            + ST layer + self-supervised prior (w/o occlusion token) &19.2M &0.49G &66.7 &43.2 &78.9 &37.1   \\
            + ST layer + self-supervised prior &19.2M &0.49G &58.2 &36.1 &67.5 &36.4   \\
            + ST layer + self-supervised prior + smoothness loss &19.2M &0.49G &58.3 &36.1 &67.1 &23.2  \\
            + ST layer + self-supervised prior + smoothness loss + gt 2D pose &19.2M &0.49G   &40.5 &23.8 &45.5 &16.3  \\
            \noalign{\hrule height 1.5pt} 
            \end{tabular}
        }
\end{center}
\vspace{-10mm}
\end{table}

\noindent\textbf{Spatial-temporal layer.}
The joint-level spatial-temporal information is essential for the occlusion problem. We first compared the temporal model without the spatial relations in~\tabref{tab:Ablation}. In addition, VIBE, TCMR, and MEVA also rely on temporal information and do not consider the kinematic features. The two comparisons in~\tabref{tab:Ablation} and~\tabref{tab:experiment} turn out the same conclusion that the model which only focuses on the temporal relation cannot achieve good results in the occluded cases. We then added a separate transformer module like~\cite{zheng20213d} in the temporal model to exploit the kinematic relation to assist the motion capture. Since the model cannot consider joint-level spatial-temporal correlation in the same stage, the performance is inferior to the spatial-temporal layer. 

\begin{figure}
    \vspace{-4mm}
    \begin{center}
    \includegraphics[width=0.5\linewidth]{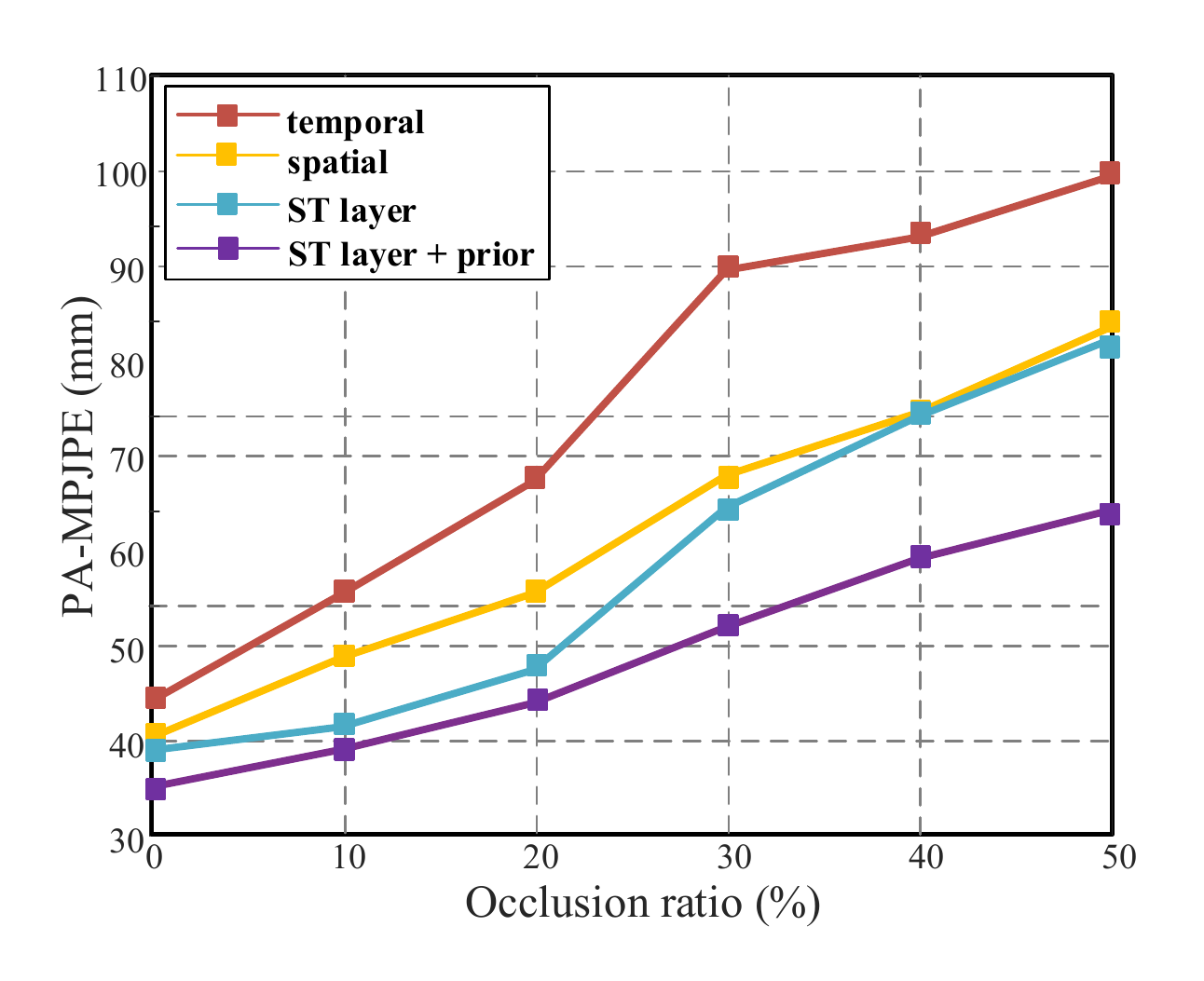}
    \end{center}
    \vspace{-8mm}
    \caption{We synthesize additional realistic occlusions on OcMotion with different occlusion ratios. The curve on different occlusion data shows that our method is more robust to variant occlusion proportions and different occlusion types.}
    \label{fig:curve}
    \vspace{-6mm}
\end{figure}

\noindent\textbf{Sensitivity to occlusions.}
We analyzed the sensitivity to the occlusion of our method to further demonstrate the effectiveness of the proposed components by synthesizing additional occlusions on OcMotion dataset. As shown in~\figref{fig:curve}, we first synthesize occluded images with different occlusion ratios to simulate severely occluded cases. Since the OcMotion dataset contains a lot of real occlusions, we synthesize realistic occlusions among the occlusion ratios from 0\% to 50\%. We use OpenPose~\cite{cao2017realtime} to detect visible 2D poses and conduct evaluations of different models on the synthetic data. The results show that the temporal approach without joint-level spatial-temporal correlation is sensitive to the occlusions, while the model with self-supervised prior is robust. With the proposed modules, our method is insensitive to various occlusions.

\section{Discussion}
\label{sec:conclusion}
In this paper, we propose a novel self-supervised prior with a joint-level spatial-temporal layer for recovering human motions from monocular occluded videos. For better generalization ability, we represent the human motion in 2D maps, and thus we can employ a lot of non-occluded 2D and 3D data for training a model that is robust to different occlusion types and various motions. The proposed method can obtain accurate and coherent motions from monocular occluded videos. To reduce the gap between synthetic and real occlusion data, we further build the first 3D occluded motion dataset~(OcMotion), which can be used for training and evaluating video-based methods for occlusion scenarios. We hope the dataset will promote future research on video-based human mesh recovery.

However, there also exist some limitations. Although the current implementation can obtain satisfactory results from a single-person occluded video, the 2D detection still affects the accuracy of 2D motion capture. When an incorrect detected joint has high confidence, it cannot be removed in the 2D occluded motion map and may cause a jittering 3D motion. The case often appears in multi-person scenarios since the inter-person occlusions are more ambiguous, and the 2D detector may predict erroneous joints with high confidence for closely interacting people. Future works may integrate more low-level vision features in the estimation and filter the noises with motion prior knowledge to prevent the undesirable impact.

\bibliographystyle{splncs04}
\bibliography{refs}

\begin{thebibliography}{10}
\providecommand{\url}[1]{\texttt{#1}}
\providecommand{\urlprefix}{URL }
\providecommand{\doi}[1]{https://doi.org/#1}

\bibitem{andriluka2018posetrack}
Andriluka, M., Iqbal, U., Insafutdinov, E., Pishchulin, L., Milan, A., Gall,
  J., Schiele, B.: Posetrack: A benchmark for human pose estimation and
  tracking. In: CVPR (2018)

\bibitem{arnab2019exploiting}
Arnab, A., Doersch, C., Zisserman, A.: Exploiting temporal context for 3d human
  pose estimation in the wild. In: CVPR (2019)

\bibitem{ba2016layer}
Ba, J.L., Kiros, J.R., Hinton, G.E.: Layer normalization. arXiv preprint
  arXiv:1607.06450  (2016)

\bibitem{bao2021beit}
Bao, H., Dong, L., Wei, F.: Beit: Bert pre-training of image transformers. In:
  ICLR (2022)

\bibitem{biggs20203d}
Biggs, B., Novotny, D., Ehrhardt, S., Joo, H., Graham, B., Vedaldi, A.: 3d
  multi-bodies: Fitting sets of plausible 3d human models to ambiguous image
  data. NeurIPS  (2020)

\bibitem{cai2019exploiting}
Cai, Y., Ge, L., Liu, J., Cai, J., Cham, T.J., Yuan, J., Thalmann, N.M.:
  Exploiting spatial-temporal relationships for 3d pose estimation via graph
  convolutional networks. In: ICCV (2019)

\bibitem{cao2017realtime}
Cao, Z., Simon, T., Wei, S.E., Sheikh, Y.: Realtime multi-person 2d pose
  estimation using part affinity fields. In: CVPR (2017)

\bibitem{chen20173d}
Chen, C.H., Ramanan, D.: 3d human pose estimation= 2d pose estimation+
  matching. In: CVPR (2017)

\bibitem{chen2021anatomy}
Chen, T., Fang, C., Shen, X., Zhu, Y., Chen, Z., Luo, J.: Anatomy-aware 3d
  human pose estimation with bone-based pose decomposition. TCSVT
  \textbf{32}(1),  198--209 (2021)

\bibitem{chen2018cascaded}
Chen, Y., Wang, Z., Peng, Y., Zhang, Z., Yu, G., Sun, J.: Cascaded pyramid
  network for multi-person pose estimation. In: CVPR (2018)

\bibitem{chen2021self}
Chen, Y.C., Piccirilli, M., Piramuthu, R., Yang, M.H.: Self-attentive 3d human
  pose and shape estimation from videos. CVIU  \textbf{213},  103305 (2021)

\bibitem{cheng2019occlusion}
Cheng, Y., Yang, B., Wang, B., Yan, W., Tan, R.T.: Occlusion-aware networks for
  3d human pose estimation in video. In: ICCV (2019)

\bibitem{choi2021beyond}
Choi, H., Moon, G., Chang, J.Y., Lee, K.M.: Beyond static features for
  temporally consistent 3d human pose and shape from a video. In: CVPR (2021)

\bibitem{choi2020pose2mesh}
Choi, H., Moon, G., Lee, K.M.: Pose2mesh: Graph convolutional network for 3d
  human pose and mesh recovery from a 2d human pose. In: ECCV (2020)

\bibitem{dosovitskiy2020image}
Dosovitskiy, A., Beyer, L., Kolesnikov, A., Weissenborn, D., Zhai, X.,
  Unterthiner, T., Dehghani, M., Minderer, M., Heigold, G., Gelly, S., et~al.:
  An image is worth 16x16 words: Transformers for image recognition at scale.
  In: ICLR (2021)

\bibitem{fang2017rmpe}
Fang, H.S., Xie, S., Tai, Y.W., Lu, C.: {RMPE}: Regional multi-person pose
  estimation. In: ICCV (2017)

\bibitem{gall2009motion}
Gall, J., Stoll, C., De~Aguiar, E., Theobalt, C., Rosenhahn, B., Seidel, H.P.:
  Motion capture using joint skeleton tracking and surface estimation. In: CVPR
  (2009)

\bibitem{he2021masked}
He, K., Chen, X., Xie, S., Li, Y., Doll{\'a}r, P., Girshick, R.: Masked
  autoencoders are scalable vision learners. In: CVPR (2022)

\bibitem{hossain2018exploiting}
Hossain, M.R.I., Little, J.J.: Exploiting temporal information for 3d human
  pose estimation. In: ECCV (2018)

\bibitem{huang2021dynamic}
Huang, B., Shu, Y., Zhang, T., Wang, Y.: Dynamic multi-person mesh recovery
  from uncalibrated multi-view cameras. In: 3DV (2021)

\bibitem{Pose2UV}
Huang, B., Zhang, T., Wang, Y.: Pose2uv: Single-shot multi-person mesh recovery
  with deep uv prior. IEEE Transactions on Image Processing  (2022)

\bibitem{huang2009estimating}
Huang, J.B., Yang, M.H.: Estimating human pose from occluded images. In: ACCV
  (2009)

\bibitem{ionescu2013human3}
Ionescu, C., Papava, D., Olaru, V., Sminchisescu, C.: Human3. 6m: Large scale
  datasets and predictive methods for 3d human sensing in natural environments.
  TPAMI  \textbf{36}(7),  1325--1339 (2013)

\bibitem{jiang20103d}
Jiang, H.: 3d human pose reconstruction using millions of exemplars. In: ICPR
  (2010)

\bibitem{johnson2010clustered}
Johnson, S., Everingham, M.: Clustered pose and nonlinear appearance models for
  human pose estimation. In: BMVC (2010)

\bibitem{joo2017panoptic}
Joo, H., Simon, T., Li, X., Liu, H., Tan, L., Gui, L., Banerjee, S., Godisart,
  T., Nabbe, B., Matthews, I., et~al.: Panoptic studio: A massively multiview
  system for social interaction capture. TPAMI  \textbf{41}(1),  190--204
  (2017)

\bibitem{kanazawa2018end}
Kanazawa, A., Black, M.J., Jacobs, D.W., Malik, J.: End-to-end recovery of
  human shape and pose. In: CVPR (2018)

\bibitem{kanazawa2019learning}
Kanazawa, A., Zhang, J.Y., Felsen, P., Malik, J.: Learning 3d human dynamics
  from video. In: CVPR (2019)

\bibitem{khirodkar2022occluded}
Khirodkar, R., Tripathi, S., Kitani, K.: Occluded human mesh recovery. In: CVPR
  (2022)

\bibitem{kocabas2020vibe}
Kocabas, M., Athanasiou, N., Black, M.J.: Vibe: Video inference for human body
  pose and shape estimation. In: CVPR (2020)

\bibitem{kocabas2021pare}
Kocabas, M., Huang, C.H.P., Hilliges, O., Black, M.J.: Pare: Part attention
  regressor for 3d human body estimation. In: ICCV (2021)

\bibitem{kolotouros2019learning}
Kolotouros, N., Pavlakos, G., Black, M.J., Daniilidis, K.: Learning to
  reconstruct 3d human pose and shape via model-fitting in the loop. In: ICCV
  (2019)

\bibitem{lee2018propagating}
Lee, K., Lee, I., Lee, S.: Propagating lstm: 3d pose estimation based on joint
  interdependency. In: ECCV (2018)

\bibitem{li2021ai}
Li, R., Yang, S., Ross, D.A., Kanazawa, A.: Ai choreographer: Music conditioned
  3d dance generation with aist++. In: ICCV (2021)

\bibitem{li2021exploiting}
Li, W., Liu, H., Ding, R., Liu, M., Wang, P., Yang, W.: Exploiting temporal
  contexts with strided transformer for 3d human pose estimation. TMM  (2022)

\bibitem{liu2019temporally}
Liu, J., Akhtar, N., Mian, A.: Temporally coherent full 3d mesh human pose
  recovery from monocular video. arXiv preprint arXiv:1906.00161  (2019)

\bibitem{liu2020attention}
Liu, R., Shen, J., Wang, H., Chen, C., Cheung, S.c., Asari, V.: Attention
  mechanism exploits temporal contexts: Real-time 3d human pose reconstruction.
  In: CVPR (2020)

\bibitem{loper2015smpl}
Loper, M., Mahmood, N., Romero, J., Pons-Moll, G., Black, M.J.: Smpl: A skinned
  multi-person linear model. TOG  \textbf{34}(6),  1--16 (2015)

\bibitem{loshchilov2017decoupled}
Loshchilov, I., Hutter, F.: Decoupled weight decay regularization. arXiv
  preprint arXiv:1711.05101  (2017)

\bibitem{luo20203d}
Luo, Z., Golestaneh, S.A., Kitani, K.M.: 3d human motion estimation via motion
  compression and refinement. In: ACCV (2020)

\bibitem{von2018recovering}
von Marcard, T., Henschel, R., Black, M.J., Rosenhahn, B., Pons-Moll, G.:
  Recovering accurate 3d human pose in the wild using imus and a moving camera.
  In: ECCV (2018)

\bibitem{martinez2017simple}
Martinez, J., Hossain, R., Romero, J., Little, J.J.: A simple yet effective
  baseline for 3d human pose estimation. In: ICCV (2017)

\bibitem{mehta2017monocular}
Mehta, D., Rhodin, H., Casas, D., Fua, P., Sotnychenko, O., Xu, W., Theobalt,
  C.: Monocular 3d human pose estimation in the wild using improved cnn
  supervision. In: 3DV (2017)

\bibitem{mehta2018single}
Mehta, D., Sotnychenko, O., Mueller, F., Xu, W., Sridhar, S., Pons-Moll, G.,
  Theobalt, C.: Single-shot multi-person 3d pose estimation from monocular rgb.
  In: 3DV (2018)

\bibitem{paszke2019pytorch}
Paszke, A., Gross, S., Massa, F., Lerer, A., Bradbury, J., Chanan, G., Killeen,
  T., Lin, Z., Gimelshein, N., Antiga, L., et~al.: Pytorch: An imperative
  style, high-performance deep learning library. NeurIPS  (2019)

\bibitem{patel2021agora}
Patel, P., Huang, C.H.P., Tesch, J., Hoffmann, D.T., Tripathi, S., Black, M.J.:
  Agora: Avatars in geography optimized for regression analysis. In: CVPR
  (2021)

\bibitem{pavllo20193d}
Pavllo, D., Feichtenhofer, C., Grangier, D., Auli, M.: 3d human pose estimation
  in video with temporal convolutions and semi-supervised training. In: CVPR
  (2019)

\bibitem{rafi2015semantic}
Rafi, U., Gall, J., Leibe, B.: A semantic occlusion model for human pose
  estimation from a single depth image. In: CVPRW (2015)

\bibitem{rempe2021humor}
Rempe, D., Birdal, T., Hertzmann, A., Yang, J., Sridhar, S., Guibas, L.J.:
  Humor: 3d human motion model for robust pose estimation. In: ICCV (2021)

\bibitem{rockwell2020full}
Rockwell, C., Fouhey, D.F.: Full-body awareness from partial observations. In:
  ECCV (2020)

\bibitem{sarandi2018robust}
S{\'a}r{\'a}ndi, I., Linder, T., Arras, K.O., Leibe, B.: How robust is 3d human
  pose estimation to occlusion? arXiv preprint arXiv:1808.09316  (2018)

\bibitem{sun2019deep}
Sun, K., Xiao, B., Liu, D., Wang, J.: Deep high-resolution representation
  learning for human pose estimation. In: CVPR (2019)

\bibitem{sun2021monocular}
Sun, Y., Bao, Q., Liu, W., Fu, Y., Black, M.J., Mei, T.: Monocular, one-stage,
  regression of multiple 3d people. In: ICCV (2021)

\bibitem{sun2021putting}
Sun, Y., Liu, W., Bao, Q., Fu, Y., Mei, T., Black, M.J.: Putting people in
  their place: Monocular regression of 3d people in depth. In: CVPR (2022)

\bibitem{sun2019human}
Sun, Y., Ye, Y., Liu, W., Gao, W., Fu, Y., Mei, T.: Human mesh recovery from
  monocular images via a skeleton-disentangled representation. In: ICCV (2019)

\bibitem{tekin2017learning}
Tekin, B., M{\'a}rquez-Neila, P., Salzmann, M., Fua, P.: Learning to fuse 2d
  and 3d image cues for monocular body pose estimation. In: ICCV (2017)

\bibitem{wan2021encoder}
Wan, Z., Li, Z., Tian, M., Liu, J., Yi, S., Li, H.: Encoder-decoder with
  multi-level attention for 3d human shape and pose estimation. In: ICCV (2021)

\bibitem{wang2020motion}
Wang, J., Yan, S., Xiong, Y., Lin, D.: Motion guided 3d pose estimation from
  videos. In: ECCV (2020)

\bibitem{wang2017outdoor}
Wang, Y., Liu, Y., Tong, X., Dai, Q., Tan, P.: Outdoor markerless motion
  capture with sparse handheld video cameras. TVCG  \textbf{24}(5),  1856--1866
  (2017)

\bibitem{wang2019geometric}
Wang, Z., Chen, L., Rathore, S., Shin, D., Fowlkes, C.: Geometric pose
  affordance: 3d human pose with scene constraints. arXiv preprint
  arXiv:1905.07718  (2019)

\bibitem{wang2022best}
Wang, Z., Yang, J., Fowlkes, C.: The best of both worlds: Combining model-based
  and nonparametric approaches for 3d human body estimation. In: CVPRW (2022)

\bibitem{xu2018monoperfcap}
Xu, W., Chatterjee, A., Zollh{\"o}fer, M., Rhodin, H., Mehta, D., Seidel, H.P.,
  Theobalt, C.: Monoperfcap: Human performance capture from monocular video.
  TOG  \textbf{37}(2),  1--15 (2018)

\bibitem{yang2022lasor}
Yang, K., Gu, R., Wang, M., Toyoura, M., Xu, G.: Lasor: Learning accurate 3d
  human pose and shape via synthetic occlusion-aware data and neural mesh
  rendering. TIP  (2022)

\bibitem{yu2020humbi}
Yu, Z., Yoon, J.S., Lee, I.K., Venkatesh, P., Park, J., Yu, J., Park, H.S.:
  Humbi: A large multiview dataset of human body expressions. In: CVPR (2020)

\bibitem{yuan2022glamr}
Yuan, Y., Iqbal, U., Molchanov, P., Kitani, K., Kautz, J.: Glamr: Global
  occlusion-aware human mesh recovery with dynamic cameras. In: CVPR (2022)

\bibitem{zhang2021learning}
Zhang, S., Zhang, Y., Bogo, F., Pollefeys, M., Tang, S.: Learning motion priors
  for 4d human body capture in 3d scenes. In: ICCV (2021)

\bibitem{zhang2020object}
Zhang, T., Huang, B., Wang, Y.: Object-occluded human shape and pose estimation
  from a single color image. In: CVPR (2020)

\bibitem{zhang2013actemes}
Zhang, W., Zhu, M., Derpanis, K.G.: From actemes to action: A
  strongly-supervised representation for detailed action understanding. In:
  ICCV (2013)

\bibitem{zhao2021travelnet}
Zhao, Z., Zhao, X., Wang, Y.: Travelnet: Self-supervised physically plausible
  hand motion learning from monocular color images. In: ICCV (2021)

\bibitem{zheng20213d}
Zheng, C., Zhu, S., Mendieta, M., Yang, T., Chen, C., Ding, Z.: 3d human pose
  estimation with spatial and temporal transformers. In: ICCV (2021)

\bibitem{zhou2017towards}
Zhou, X., Huang, Q., Sun, X., Xue, X., Wei, Y.: Towards 3d human pose
  estimation in the wild: a weakly-supervised approach. In: ICCV (2017)

\bibitem{zhou2019continuity}
Zhou, Y., Barnes, C., Lu, J., Yang, J., Li, H.: On the continuity of rotation
  representations in neural networks. In: CVPR (2019)

\end{thebibliography}

\end{document}